\pdfoutput=1

\documentclass[11pt]{article}

\usepackage[]{acl}

\usepackage{times}
\usepackage{latexsym}

\usepackage[T1]{fontenc}

\usepackage[utf8]{inputenc}

\usepackage{microtype}

\usepackage{graphicx}
\usepackage{amsmath}
\usepackage{booktabs}
\usepackage{multirow}
\usepackage{array}
\usepackage{threeparttable}
\usepackage{rotating}
\usepackage{cleveref}

\crefname{section}{§}{§§}
\Crefname{section}{§}{§§}
\title{TIE: Topological Information Enhanced Structural Reading Comprehension on Web Pages}

\author{Zihan Zhao, Lu Chen\footnotemark[1], Ruisheng Cao, Hongshen Xu, Xingyu Chen \and Kai Yu\footnotemark[1]\\
  X-LANCE Lab, Department of Computer Science and Engineering \\
  MoE Key Lab of Artificial Intelligence, AI Institute, Shanghai Jiao Tong University, China \\
  Shanghai Jiao Tong University, Shanghai, China \\
  State Key Laboratory of Media Convergence Production Technology and Systems \\
  \texttt{zhao\_mengxin@sjtu.edu.cn, chenlusz@sjtu.edu.cn}\\
  \texttt{\{211314, xuhongshen, galaxychen, kai.yu\}@sjtu.edu.cn} \\}

\begin{document}
\maketitle
\renewcommand{\thefootnote}{\fnsymbol{footnote}}
\footnotetext[1]{The corresponding authors are Lu Chen and Kai Yu.}
\renewcommand{\thefootnote}{\arabic{footnote}}
\begin{abstract}

Recently, the structural reading comprehension (SRC)  task on web pages has attracted increasing research interests. Although previous SRC work has leveraged extra information such as HTML tags or XPaths, the informative topology of web pages is not effectively exploited. In this work, we propose a \textbf{T}opological \textbf{I}nformation \textbf{E}nhanced model~(TIE), which transforms the token-level task into a tag-level task by introducing a two-stage process (i.e. \textit{node locating} and \textit{answer refining}). Based on that, TIE integrates Graph Attention Network~(GAT) and Pre-trained Language Model~(PLM) to leverage the topological information of both logical structures and spatial structures. Experimental results demonstrate that our model outperforms strong baselines and achieves state-of-the-art performances on the web-based SRC benchmark WebSRC at the time of writing. The code of TIE will be publicly available at \url{https://github.com/X-LANCE/TIE}.

\end{abstract}
\section{Introduction}\label{sec:intro}

With the rapid development of the Internet, web pages have become the most common and rich source of information~\citep{web}.
Therefore, the ability to understand the contents of structured web pages will guarantee a rich and diverse knowledge source for deep learning systems.
Each web page is mainly rendered from the corresponding HyperText Markup Language~(HTML) codes. In other words, the understanding of a structured web page can be achieved by the comprehension of its HTML codes.

\begin{figure}[t]
    \centering
    \includegraphics[width=\hsize]{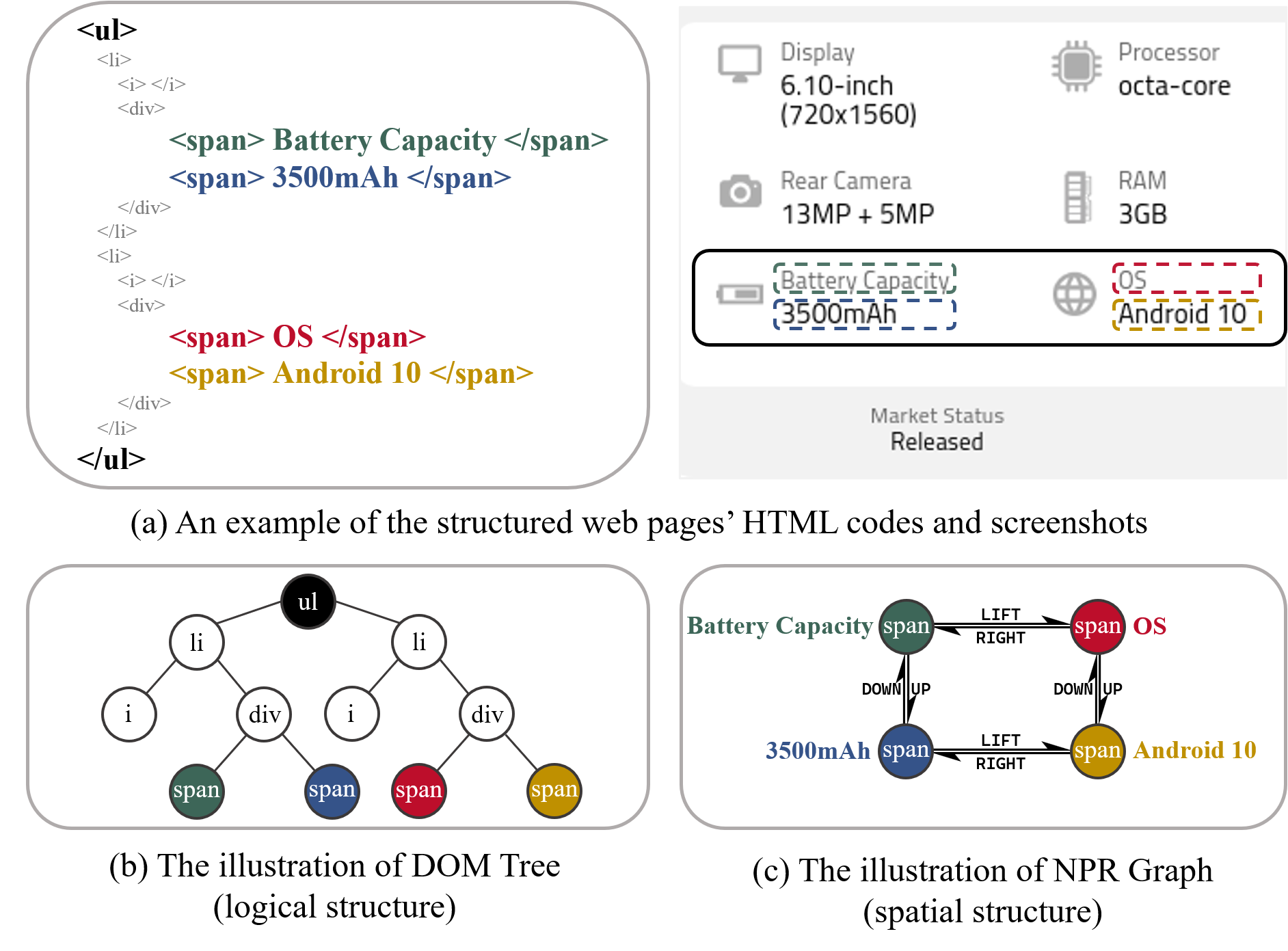}
    \caption{An example of web pages in WebSRC and its corresponding Document Object Model~(DOM) tree and Node Positional Relation~(NPR) graph in WebSRC. The colored HTML tag in (a) is corresponding to the bounding box with the same color in (a) and the node with the same color in (b) and (c).}
    \label{fig:data}
\end{figure}

One of the commonly used tasks to verify the model's ability of comprehension is Question Answering~(QA). However, previous QA models only focus on the comprehension of plain texts \citep{rajpurkar-etal-2016-squad, yang-etal-2018-hotpotqa, reddy-etal-2019-coqa, mrc}, tables~\citep{pasupat-liang-2015-compositional, chen-etal-2020-hybridqa, ttqa}, or knowledge bases~(KBs)~\citep{berant-etal-2013-semantic, talmor-berant-2018-web}.
These sources have either no topological structure or fixed-form structures.
On the contrary, the topological structures of web pages are complex and flexible, which are less investigated in previous QA works.

Specifically, HTML codes can be viewed as multiple semantic unit separated by tag tokens (e.g. {\tt <div>}, {\tt </div>}). An HTML tag refers to a pair of matched start and end tags and all the content in between, which also corresponds to a part of the web page (illustrated in Fig. \ref{fig:data} (a)). Therefore, there are two kinds of topological structures in web pages: logical structures which contain the hierarchical relations and clustering of tags (see Fig. \ref{fig:data} (b)); and spatial structures which contain the relative positions between different tags in the web pages (see Fig. \ref{fig:data} (c)). These topological structures are as important as the semantics of HTML codes and screenshots.

Although previous works~\citep{websrc, markuplm} have tried to leverage the topological structures by adopting HTML tags or XPaths as tokens or position embeddings, only logical structures are encoded implicitly. However, it is obvious for humans to identify key-value pairs if two spans are located in the same row or column, while this relation may take various forms in the logical structures of different web pages. Moreover, tables have extremely simple spatial structures but will be super complex in terms of logical structures. Therefore, spatial structures are essential and complementary to logical structures.

The major obstacle that prevents previous models to leverage spatial relations is that both the two kinds of topological structures are organized at the tag level instead of the token level (Fig. \ref{fig:data} (b) and (c)). As token-level models, whose computation and prediction units are the tokens of web pages, it is extremely hard and anti-natural for them to encode the topological structures.
Moreover, using token-level models also means that previous works have to implicitly imply the logical structures to the models, which may be less effective than explicitly telling with the help of prior knowledge. 

To tackle these problems, we propose
\textbf{T}opological \textbf{I}nformation \textbf{E}nhanced model~(TIE), a tag-level QA model that operates on the representations of HTML tags to predict which tag the answer span belongs to.
By switching from token level to tag level,
various structures of web pages can be explicitly encoded into the model easily.
Specifically, TIE encodes both the logical and spatial structures using Graph Attention Network~(GAT) \citep{gat} with the help of two kinds of graphs. The first kind of graphs is \textit{Document Object Model~(DOM) trees} which is widely used to represent the logical structures of HTML codes.
Secondly, to encode the spatial structures,
we define the \textit{Node Positional Relation~(NPR) graph} based on the bounding box of HTML tags obtained by the browser. Detail definition can be found in Section \ref{sec:npr}.

Moreover, to accomplish the token-level prediction tasks by a tag-level QA model, we further introduce a two-stage process including \textit{node locating} stage and \textit{answer refining} stage. Specifically, in the \textit{answer refining} stage, a traditional token-level QA model is utilized to extract answer span with the constraint of the answer node prediction by TIE in the \textit{node locating} stage.

Our TIE model is tested on the WebSRC dataset \footnote{\url{https://x-lance.github.io/WebSRC/}.} and achieve state-of-the-art~(SOTA) performances.

To summarize, our contributions are three folds:
\begin{itemize}
    \item We propose a tag-level QA model called TIE with a two-stage inference process: \textit{node locating} stage and \textit{answer refining} stage.
    \item We utilize GAT to leverage the topological information of both the logical and spatial structures with the help of DOM trees and our newly defined NPR graphs.
    \item Experimental results on the WebSRC dataset demonstrate the effectiveness of our model and its key component.
\end{itemize}

\section{Preliminary}

\subsection{Task Definition}\label{sec:td}

The Web-based SRC task \citep{websrc} is defined as a typical extractive question answering task based on web pages. Given the user query $\boldsymbol{q}=(q_1, q_2,\cdots, q_{|\boldsymbol{q}|})$ and the flattened HTML code sequence $\boldsymbol{c}=(c_1, c_2,\cdots, c_{|\boldsymbol{c}|})$ of relevant web page as inputs , the goal is to predict the starting and ending position of answer span $(s, e)$
in the HTML codes $\boldsymbol{c}$ where $|\boldsymbol{q}|, |\boldsymbol{c}|$
denote the length of the question and the HTML code sequence, respectively,
and $1 \leq s \leq e \leq |\boldsymbol{c}|$. Notice that each token $c_i$ in the flattened HTML codes $\boldsymbol{c}$ can be a raw text word or tag symbol such as {\tt <div>} while the user query $q$ is a word sequence of plain text.

\begin{figure}[t]
    \centering
    \includegraphics[width=\hsize]{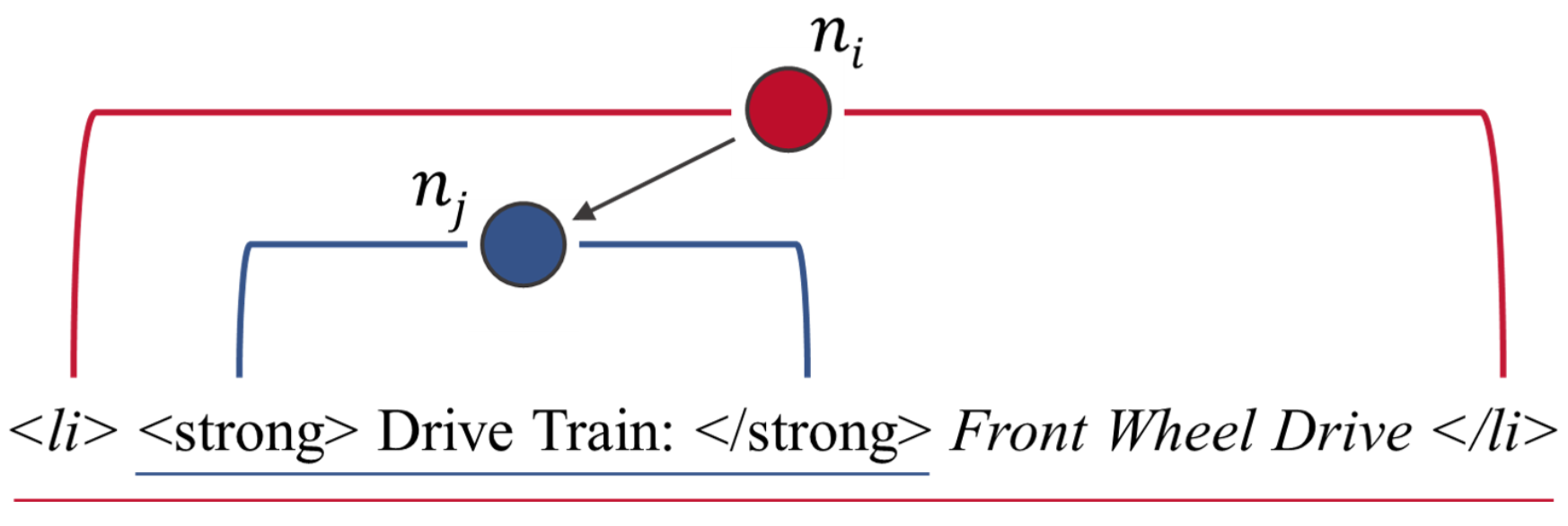}
    \caption{Illustration of the relations between DOM trees and HTML codes. The italic tokens "<li> Front Wheel Drive </li>" are the \textit{direct content} of node $n_i$}
    \label{fig:dom}
\end{figure}

\subsection{DOM Trees of HTML codes}\label{sec:dom}

The DOM tree is a special tree structure that is parsed from raw HTML codes by Document Object Model~\footnote{\url{https://en.wikipedia.org/wiki/Document_Object_Model}}. Each node in the tree denotes a tag closure in the original HTML code. Specifically, each node contains a start tag token~(e.g. {\tt <div>}), an end tag token~(e.g. {\tt </div>}), and all the contents in between. One DOM node $n_j$ is the descendant of another node $n_i$, iff the contents of node $n_j$ is entirely included in the contents of node $n_i$. 

Furthermore, we define the \textit{direct contents} of each DOM node (and its corresponding HTML tag) as all the tokens in its tag closure
that are not contained in any of its children (see Figure \ref{fig:dom}).
\section{TIE}

In this section, we will first introduce the architecture of the whole SRC system in Sec.\ref{sec:inf}, and then the two kind of graph we used in Sec. \ref{sec:graph}. Finally, the structure of \textbf{T}opological \textbf{I}nformation \textbf{E}nhance model~(TIE) is demonstrated in Sec.\ref{sec:model}. 

\subsection{Architecture of the Whole SRC System}\label{sec:inf}

\begin{figure}[t]
    \centering
    \includegraphics[width=\hsize]{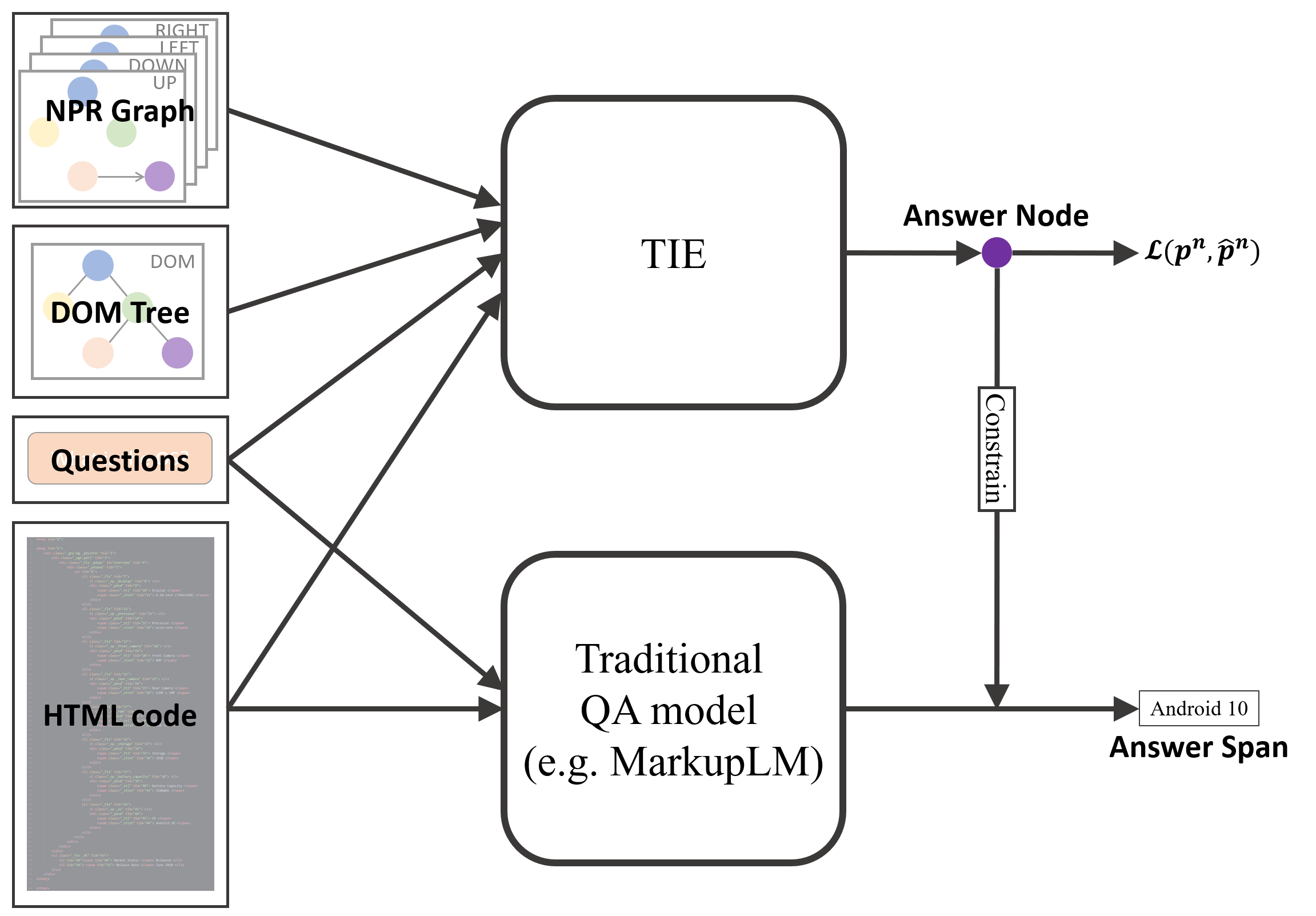}
    \caption{The two-stage architecture using TIE and traditional QA model (e.g. MarkupLM)}
    \label{fig:ppl}
\end{figure}

With the help of DOM trees and NPR graphs, TIE can efficiently predict in which node the answer is located. Therefore,
we modify the original architecture of the SRC system into a two-stage architecture: \textit{node locating} and \textit{answer refining}.
The two-stage architecture is illustrated in Figure \ref{fig:ppl}.

In the \textit{node locating} stage, we first define the answer node as the deepest node in the DOM tree which contains the complete answer span. Then, TIE is utilized to predict the answer node $n_a$ for the question $q$ given the flattened HTML codes $c$ and the corresponding DOM tree $\mathcal{D}_{c}$ and NPR graphs $\mathcal{G}_{c}$~(see Sec. \ref{sec:graph}). Formally,
\begin{gather*}
    \text{TIE}(q, c, (\mathcal{D}_{c}, \mathcal{G}_{c})) = \boldsymbol{p}^n,\\
    n_a = \underset{n_i\in V_{D_c}}{\text{argmax}}(p_i^n),
\end{gather*}
where $p_i^n$ denotes the probability of node $n_i$ being the answer node, and $V_{D_c}$ is the node set of $D_c$.

Then, in the \textit{answer refining} stage, we use the predicted answer node as a constraint during the prediction of the answer span.
In more detail, we first use a QA model (e.g. MarkupLM) to obtain the start and end probabilities $\boldsymbol{p}^s$, $\boldsymbol{p}^e$ among all the tokens of HTML code sequence $c$.
Then, the predicted answer span is chosen from the spans which are contained by the predicted answer node $n_a$. 
To conclude, provided that the starting and ending position of predicted answer node $n_a$ in the HTML code $c$ is $s_a$, and $e_a$, the second stage can be formulated as follows:
\begin{gather*}
    \text{QA}(q, c) = \boldsymbol{p}^s, \boldsymbol{p}^e \\
    (s_{pred}, e_{pred}) = \underset{(i, j): s_a\leq i<j\leq e_a}{\text{argmax}}(p^s_i + p^e_j)
\end{gather*}

\begin{figure*}[t]
    \centering
    \includegraphics[width=\textwidth]{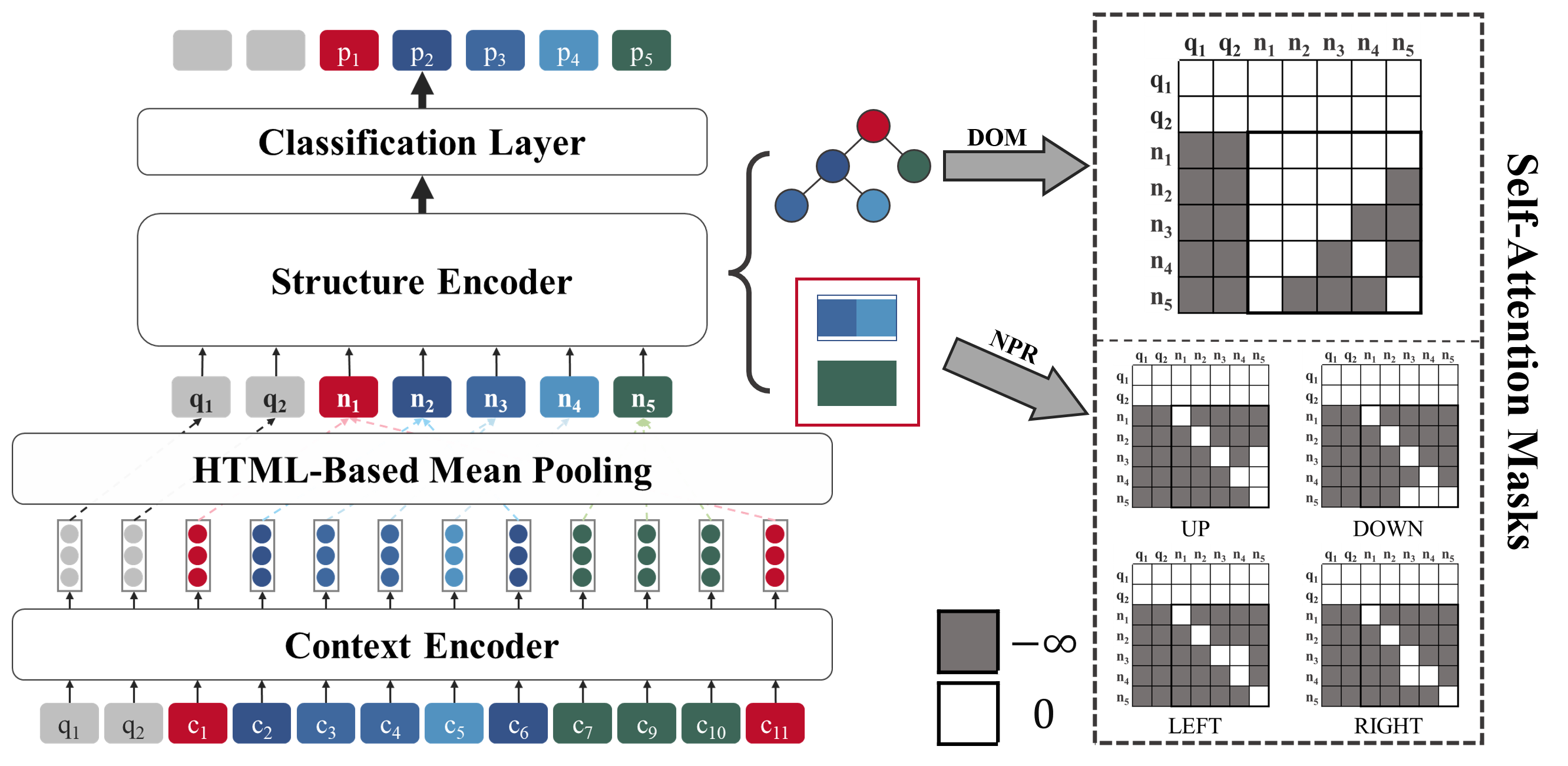}
    \caption{The overall architecture of TIE}
    \label{fig:model}
\end{figure*}

\subsection{Construction of GAT Graphs}\label{sec:graph}

Recently, Graph Neural Network~(GNN) \citep{scarselli2008graph} has been widely used in multiple Neural Language Processing tasks, such as text classification and generation \citep{yao2019graph, zhao-etal-2020-line}, information extraction \citep{lockard-etal-2020-zeroshotceres}, dialogue policy optimization \citep{chen2018policy, chen-etal-2018-structured, chen2019agentgraph, chen2020distributed},  dialogue state tracking \citep{chen2020schema, zhu-etal-2020-efficient}, Chinese processing \citep{gui-etal-2019-lexicon, chen-etal-2020-neural-graph, lyu2021let}, etc.
Graph Attention Network~(GAT) is a special type of GNN that encodes graphs with attention mechanism. In this work, to leverage both the logical and spatial structures, we introduce two kinds of graphs: DOM Trees and NPR graphs.

\subsubsection{DOM Trees}
The logical relations of HTML codes can be described with the assistance of its DOM Tree~(see Sec. \ref{sec:dom}). However, the original tree is extremely sparse, which often leads to poor communication efficiency among nodes.
To this end, we modify the structure to enlarge the receptive fields for each node. Mathematically, the resulting graph $\mathcal{D}_{c}=(V_{D_c}, E_{D_c})$ can be constructed from the original sparse form $\mathcal{D}=(V_D,E_D)$,
\begin{align*}
\begin{cases}
 \begin{split}
 V_D =& \text{all nodes in the original DOM tree,} \\
 E_D =& \{(n_i,n_j)|n_i\text{ is the parent of }n_j\}\cup\\
     &\{(n_i,n_j)|n_i\text{ is a child of }n_j\},
 \end{split}
\end{cases}
\end{align*}
into a denser one $\mathcal{D}_{c}=(V_{D_c},E_{D_c})$,
\begin{align*}
\begin{cases}
     \begin{split}
     V_{D_c} =& V_D \\
     E_{D_c} =& \{(n_i, n_i)| n_i\in V_{D_c}\}\cup\\
         &\{(n_i,n_j)|n_i\text{ is an {\bf ancestor} of }n_j\}\cup\\
         &\{(n_i,n_j)|n_i\text{ is a {\bf descendant} of }n_j\}
     \end{split}
\end{cases}
\end{align*}
In this way, each node can directly communicate with all of its ancestors and descendants, so that the information can be transferred much faster.

\subsubsection{NPR Graphs}\label{sec:npr}

To explicitly establish the positional relations between different texts, we define and construct Node Positional Relation~(NPR) graph $\mathcal{G}_c=(V_G, E_G)$ based on the rendered structured web pages.

Similar to DOM Tree, each NPR node $n_i$ corresponds to a tag $t_i$ in the HTML code of the web page. The content of NPR nodes is defined as the direct content of their corresponding HTML tags. It is worth noticing that under our definition, the node sets of the NPR graph and the DOM tree of the same web page are identical ($V_G=V_D$).

Moreover, considering that the nodes with informative relations (such as {\tt "key-value"} relations and {\tt "header-cell"} relations) are usually located on the same row or column, we introduce four kinds of directed edges into NPR graphs: {\tt UP}, {\tt DOWN}, {\tt LEFT}, and {\tt RIGHT}. Specifically, $(n_i, n_j) \in E_G^{\tt UP}$ when
\begin{equation}
\left\{
\begin{aligned}
\min(x_{n_i} + &w_{n_i}, x_{n_j} + w_{n_j}) - \max(x_{n_i}, x_{n_j}) \\
&\geq \gamma \times \min(w_{n_i}, w_{n_j})\label{for:npr1}\\
y_{n_i} \geq\ &y_{n_j}\ \text{or}\ y_{n_i}+h_{h_i} \geq y_{n_j}+h_{n_j} \\
\end{aligned}
\right.
\end{equation}
both hold, where ($x_{n_i}$, $y_{n_i}$), ($x_{n_j}$, $y_{n_j}$) are the coordinates of the upper-left corner of the bounding boxes corresponding to the nodes $n_i$ and $n_j$; $w_{n_i}$, $w_{n_j}$ are the width of the two bounding boxes while $h_{n_i}$, $h_{n_j}$ are the height of the two bounding boxes; and $\gamma$ is a hyper-parameter. Similar functions are used for $E_G^{\tt DOWN}$, $E_G^{\tt LEFT}$, and $E_G^{\tt RIGHT}$. Finally, $E_G = E_G^{\tt UP}\bigcup E_G^{\tt DOWN} \bigcup E_G^{\tt LEFT} \bigcup E_G^{\tt RIGHT}$
Figure \ref{fig:data} (a) and (c) show an example of the NPR graph and its corresponding HTML code.

To simplify the NPR graphs, we only consider the nodes whose direct contents contain text tokens. That means in NPR graphs, the nodes whose direct contents only contain tag tokens will be isolated nodes with no relation.

\subsection{Design of TIE}\label{sec:model}

The model we proposed, TIE, mainly consists of four parts: the \textbf{Context Encoder Module}, the \textbf{HTML-Based Mean Pooling}, the \textbf{Structure Encoder Module}, and the \textbf{Classification Layer}. The overall architecture of TIE is shown in Figure \ref{fig:model}.

\paragraph{Context Encoder Module.} We first utilize Pre-trained Language Model as our context encoder. It encodes the contextual information of the HTML codes and gets the contextual word embeddings used for node representation initialization.
Specifically, we use two PLM in our experiments:
H-PLM~\citep{websrc} + RoBERTa~\citep{roberta} and MarkupLM~\citep{markuplm}.

\paragraph{HTML-Based Mean Pooling.} In this module, TIE initializes the node representations based on the contextual word embedding calculated by Context Encoder.
Specifically, for each node, we initialize its representation as the average embedding of its corresponding tag's direct contents.
Formally, the representation of node $n_i$ is calculated as:
\begin{equation}
\boldsymbol{n_i} = \underset{x_j\in\text{DC}(n_i)}{mean}(\boldsymbol{x}_j)
\end{equation}
where $\text{DC}(n_i)$ means the tokens set of the direct contents of node $n_i$; $\boldsymbol{x}_j$ is the contextual embedding of token $x_j$. 

\paragraph{Structure Encoder Module.} TIE utilizes GAT to encode the topological information contained in DOM trees and NPR graphs.
Specifically, for the i-th attention head of GAT:
\begin{gather*}
    \boldsymbol{Q}_i = \boldsymbol{W}_{q,i}\boldsymbol{N};\ 
    \boldsymbol{K}_i = \boldsymbol{W}_{k,i}\boldsymbol{N};\ 
    \boldsymbol{V}_i = \boldsymbol{W}_{v,i}\boldsymbol{N}\\
    \text{GAT}_i(\boldsymbol{N})=\text{softmax}(\frac{\boldsymbol{Q_i}^T\boldsymbol{K_i}}{\sqrt{d}} + \boldsymbol{M_i})\boldsymbol{V_i}\\
    m^{(i)}_{jk}=
    \left\{
    \begin{array}{ll}
    0 & (n_j, n_k) \in \text{Edge}(G_i)\\
    - \infty & otherwise
    \end{array}
    \right.\\
    G_i\in \{\mathcal{D}_{c}, \mathcal{G}^{\tt UP}_c, \mathcal{G}^{\tt DOWN}_c, \mathcal{G}^{\tt LEFT}_c, \mathcal{G}^{\tt RIGHT}_c\}
\end{gather*}
where $\boldsymbol{N}=[\boldsymbol{n}_i]_{d\times |\mathcal{N}|}$; $d$ is the dimension of the node representations $\boldsymbol{n}_i$; $\boldsymbol{W}_i$s are the learnable parameters; $\boldsymbol{M}_i=[m^{(i)}_{jk}]_{|\mathcal{N}|\times |\mathcal{N}|}$ is the mask matrix for the i-th attention head. Finally, the outputs of all the attention heads are concatenated to form the node representations for the next GAT layer.

\paragraph{Classification Layers.} Finally, we get the embeddings of all the nodes from the Structure Encoder Module and utilize a single linear layer followed by a Softmax function to calculate each node's probability of being the answer node.

\section{Experiments}

\subsection{Dataset}

\begin{table}[t]
    \centering
    \begin{tabular}{c|c|c|c|c}
         \toprule
         \multirow{2}{*}{Type} & \multicolumn{2}{c|}{Training set} & \multicolumn{2}{c}{Dev set} \\
         \cline{2-5}
          & \#QA & \% & \#QA & \% \\
         \midrule
         KV & 129990 & 42.3 & 21798 & 41.3 \\
         Comparison & 52893 & 12.2 & 9078 & 17.2 \\
         Table & 124432 & 40.5 & 21950 & 41.6 \\
         \bottomrule
    \end{tabular}
    \caption{The statistics of QA pairs from different types of websites in WebSRC.}
    \label{tab:dataset}
\end{table}

We evaluate our proposed methods on WebSRC \citep{websrc}.
In more detail, the WebSRC dataset consists of 0.4M question-answer pairs and 6.4K web page segments with complex structures. For each web page segment, apart from its corresponding HTML codes, the dataset also provides the bounding box information of each HTML tag obtained from the rendered web page. Therefore, we can easily use this information to construct the NPR graph for each web page segment.

Moreover, WebSRC groups the websites into three classes: \textit{KV}, \textit{Comparison}, and \textit{Table}. Specifically, \textit{KV} indicates that the information in the websites is mainly presented in the form of {\tt "key:value"}, where {\tt key} is an attribute name and {\tt value} is the corresponding value. \textit{Comparison} indicates that each web page segment of the websites contains several entities with the same set of attributes. \textit{Table} indicates that the websites mainly use a table to present information. The statistics of different types of websites in WebSRC are shown in Table \ref{tab:dataset}.

We submit our models to the official of WebSRC for testing.

\subsection{Metrics}

\begin{table*}[t]
    \centering
    \begin{tabular}{p{0.015\textwidth}<{\centering}|p{0.35\textwidth}<{\centering}|p{0.07\textwidth}<{\centering}p{0.07\textwidth}<{\centering}p{0.07\textwidth}<{\centering}|p{0.07\textwidth}<{\centering}p{0.07\textwidth}<{\centering}p{0.07\textwidth}<{\centering}}
         \toprule
         &\multirow{2}{*}{Method} & \multicolumn{3}{c|}{Dev} & \multicolumn{3}{c}{Test} \\
         \cline{3-8}
         && EM$\uparrow$ & F1$\uparrow$ & POS$\uparrow$ & EM$\uparrow$ & F1$\uparrow$ & POS$\uparrow$ \\
         \hline
         \hline
         \multirow{6}{*}{\begin{sideways}BASE\end{sideways}} & T-PLM(BERT)~\citep{websrc} & 52.12 & 61.57 & 79.74 & 39.28 & 49.49 & 67.68 \\
         &H-PLM(BERT)~\citep{websrc} & 61.51 & 67.04 & 82.97 & 52.61 & 59.88 & 76.13 \\
         &V-PLM(BERT)~\citep{websrc} & 62.07 & 66.66 & 83.64 & 52.84 & 60.80 & 76.39 \\
         &$\text{MarkupLM}$~\citep{markuplm} & 68.39 & 74.47 & 87.93 & - & - & - \\
         \cline{2-8}
         & $\text{MarkupLM}^{*}$ & 68.99 & 74.55 & 88.40 & 60.43 & 67.05 & 80.55 \\
         & $\text{TIE}_{\text{MarkupLM}}$ & 76.83 & 82.77 & 90.90 & 71.86 & 75.91 & 85.74 \\
         \midrule
         \multirow{8}{*}{\begin{sideways}LARGE\end{sideways}} & T-PLM(Electra)~\citep{websrc} & 61.67 & 69.85 & 84.15 & 56.32 & 72.35 & 79.18 \\
         &H-PLM(Electra)~\citep{websrc} & 70.12 & 74.14 & 86.33 & 66.29 & 72.71 & 83.17 \\
         &V-PLM(Electra)~\citep{websrc} & 73.22 & 76.16 & 87.06 & 68.07 & 75.25 & 84.96 \\
         &$\text{MarkupLM}$~\citep{markuplm} & 74.43 & 80.54 & 90.15 & - & - & - \\
         \cline{2-8}
         &$\text{H-PLM(RoBERTa)}^*$ & 70.90 & 75.15 & 87.16 & 67.76 & 74.61 & 86.29 \\
         &$\text{TIE}_{\text{H-PLM(RoBERTa)}}$ & 75.57 & 79.38 & 88.29 & 69.65 & 74.78 & 85.72 \\
         \cline{2-8}
         &$\text{MarkupLM}^{*\dag}$ & 73.38 & 79.83 & 89.93 & 69.09 & 76.45 & 87.24 \\
         &$\text{TIE}_{\text{MarkupLM}} ^{\dag}$\tnote{2} & \textbf{81.66} & \textbf{86.24} & \textbf{92.29} & \textbf{75.87} & \textbf{80.19} & \textbf{89.73} \\
         \bottomrule
    \end{tabular}
    \caption{The results of our proposed method on WebSRC. EM denotes the exact match scores; F1 denotes the token level F1 scores; POS denotes the path overlap scores. We submit the models to the official of WebSRC for testing. * denotes reproduction results. \dag denotes average results of 3 random seeds.}
    \label{tab:result}
\end{table*}

To keep consistent with previous studies, we adopt the following three metrics: (1) Exact Match~(EM), which measures whether the predicted answer span is exactly the same as the golden answer span. (2) Token level F1 score~(F1), which measures the token level overlap of the predicted answer span and the golden answer span. (3) Path Overlap Score~(POS), which measures the overlap of the path from the root tag~({\tt <HTML>}) to the deepest tag that contains the complete predicted answer span and that contains the complete golden answer span. Formally, the POS is calculated as follows:
\begin{equation}
POS=\frac{|P_{pred} \bigcap P_{gt}|}{|P_{pred} \bigcup P_{gt}|} \times 100\%
\end{equation}
where $P_{pred}$ and $P_{gt}$ are the set of tags that on the path from the root~({\tt <HTML>}) tag to the deepest tag that contains the complete predicted answer span or the ground truth answer span, respectively.

\subsection{Baselines \& Setup}

We leverage the three models introduced in \citet{websrc} and MarkupLM \citep{markuplm} as our baselines. Specifically, T-PLM converts the HTML codes into plain text by simply removing all the HTML tags, while H-PLM treats HTML tags as special tokens and uses the origin HTML code sequences as input. Then, both of them utilize PLMs to generate the predicted answer span. To leverage visual information, V-PLM concatenates token embeddings resulting from H-PLM with visual embeddings and then feeds the results into multiple self-attention blocks before generating predictions. Faster R-CNN is utilized to extract visual embeddings from screenshots of the corresponding web pages. On the other hand, MarkupLM leverages XPaths to encode the logical position of each token and use it as an additional position embedding.

In our experiments, we use 3 GAT blocks as the Structure Encoder Module of TIE. H-PLM(RoBERTa) and MarkupLM are leveraged as context encoders. The implementation of TIE is based on the official code provided by WebSRC~\footnote{\url{https://github.com/X-LANCE/WebSRC-Baseline}} and MarkupLM~\footnote{\url{https://github.com/microsoft/unilm/tree/master/markuplm}}.
We set the hyperparameter
$\gamma$ in Eq.\ref{for:npr1}
to be 0.5.
Finally, the models used in the \textit{answer refining} stage are of the same architecture as the context encoder models of TIE while individually trained on WebSRC.
For more setup details, please refer to Appendix. \ref{sec:setup}

\subsection{Main Results}\label{sec:result}

The experimental results on the development set and the test set are shown in Table \ref{tab:result}. Specifically, the performances of TIE in the following sections
refer to the performances of the proposed two-stage system, and the subscript of TIE refer to both the context-encoder for TIE and the QA model used in \textit{answer refining} stage.

\begin{figure*}[t]
    \centering
    \includegraphics[width=\textwidth]{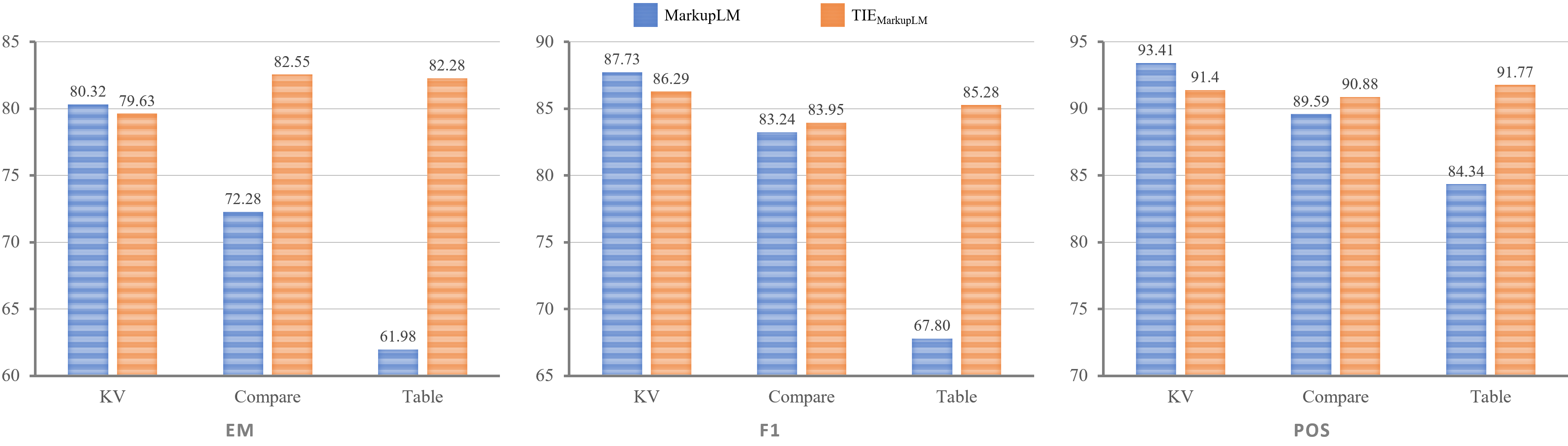}
    \caption{The performance comparison on different types of websites of the development set.}
    \label{fig:exp}
\end{figure*}

From the results, we can find out that our TIE consistently achieves better results compared with the corresponding baselines.
Specifically, $\text{TIE}_{\text{MarkupLM}}$ significantly outperforms the previous SOTA results, MarkupLM, by 6.78\% EM, 3.74\% F1, and 2.49\% POS on the test set. Moreover, it is worth noticing that the performance of $\text{TIE}_\text{MarkupLM-BASE}$ is even higher than the performance of the MarkupLM-LARGE model (76.83\% v.s. 73.38\% EM on the development set and 71.86\% v.s. 69.09\% EM on the test set).
These results strongly demonstrate that TIE can effectively model the topological information of the semi-structured web pages with the help of its structure encoder.

\begin{table}[tb]
    \centering
    \begin{tabular}{c|p{0.06\textwidth}<{\centering}|p{0.06\textwidth}<{\centering}|c}
        \toprule
          & $|S_0|$ & $|S_1|$ & $|S_0|:|S_1|$ \\
         \midrule
         MarkupLM & 873 & 692 & 1.26:1 \\
         $\text{TIE}_{\text{MarkupLM}}$ & 944 & 314 & 3.1:1 \\
         \bottomrule
    \end{tabular}
    \caption{The statistics of samples on \textit{Compare} websites in the development set with wrong predictions. $S_0$ is the set of examples with 0 F1 scores. $S_1$ is the set of examples with F1 scores between 0 and 1. The numbers are average results of 3 random seeds.}
    \label{tab:f1}
\end{table}

Furthermore, we compare the performances of $\text{TIE}_{\text{MarkupLM}}$ and MarkupLM on different types of websites. The results are shown in Figure \ref{fig:exp}. From the figure, we find that our method achieves significant improvements on the websites of type \textit{Table} (+20.30\% EM, +17.48\% F1, +7.43\% POS) while suffering slight performance drops on the websites of type \textit{KV}. We hypothesize the reason is that topological structures are less important in the websites of type \textit{KV}, so that stronger contextual encoding abilities will lead to better results. More analysis can be found in Sec. \ref{sec:case}.

We also notice that the improvements of F1 are less considerable compared with those of EM on the websites of type \textit{Compare} (+10.27\% EM v.s. +0.71\% F1). The reason lies in the cascading error of our two-stage process. Specifically, in the \textit{node locating} stage, the model may generate a wrong prediction which is not one of the ancestors of the answer node. In this case, as the answer span is not contained in the predicted node, the final F1 score is highly likely to be zero. Detailed calculations, see Table \ref{tab:f1}, strongly support our analysis.

\subsection{Case Study}\label{sec:case}

\begin{figure*}[t]
    \centering
    \includegraphics[width=\hsize]{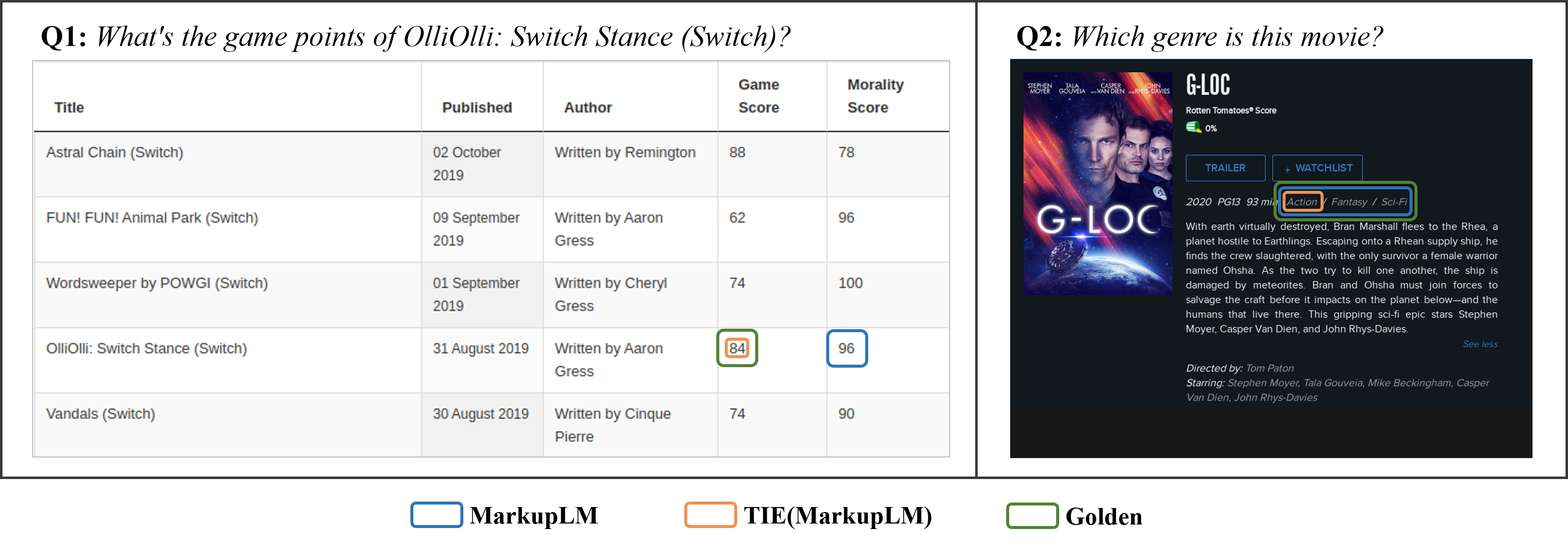}
    \caption{Examples of the results in the development set.}
    \label{fig:com}
\end{figure*}

In Fig. \ref{fig:com}, we compare the answers generated by $\text{TIE}_{\text{MarkupLM}}$ and MarkupLM. More examples can be found in Appendix. \ref{sec:add}.

\textbf{Q1} is a typical example of \textit{Table} websites. It is obvious that multiple {\tt "header-cell"} relations need to be recognized when answering \textbf{Q1}. Specifically, one should first find "\textit{OlliOlli: Switch Stance (Switch)}" from column "\textit{Title}" (first {\tt "header-cell"} relation), then locate the answer at the crossing cell of row "\textit{OlliOlli: Switch Stance (Switch)}" (second {\tt "header-cell"} relation) and column "\textit{Game Score}" (third {\tt "header-cell"} relation). With the help of topological information, TIE can correctly answer this question. However, MarkupLM only successfully locates the row and fails to recognize the long range relation between "\textit{Game Score}" and "\textit{84}". Considering that this row can also be identified by string matching, this example strongly demonstrate that TIE is much stronger in terms of long range topological relation encoding.

\textbf{Q2} is a typical example of \textit{KV} websites. The topological structures of this web page are far less complex. To answer \textbf{Q2}, the most important step is to discover the semantic similarity among "\textit{Action}", "\textit{Fantasy}", and "\textit{Sci-Fi}" and then group them together. In this case, the contextual distances of these words will be extremely helpful. Therefore, MarkupLM is able to generate the correct prediction. However, as TIE focuses on the comprehension of node structures where sequencing order and semantics are less valuable, TIE fails to group the three nodes.

\subsection{Ablation Study}\label{sec:abl}

\begin{table}[t]
    \centering
    \small
    \begin{tabular}{p{0.3cm}<{\centering}@{-}l|ccc}
         \toprule
         \multicolumn{2}{c|}{Method} & EM$\uparrow$ & F1$\uparrow$ & POS$\uparrow$ \\
         \hline
         \hline
         \multicolumn{2}{c|}{$\text{TIE}_{\text{MarkupLM}}^{\dag}$} & 81.66 & 86.24 & 92.29 \\
         \hline
         & w/o $\text{DOM}^{\dag}$ & $\text{81.05}_{\text{(-0.61)}}$ & $\text{85.42}_{\text{(-0.82)}}$ & $\text{91.62}_{\text{(-0.67)}}$ \\
         & w/ ORD & $\text{72.20}_{\text{(-9.46)}}$ & $\text{77.80}_{\text{(-8.44)}}$ & $\text{89.39}_{\text{(-1.90)}}$ \\
         \hline
         & w/o NPR & $\text{72.62}_{\text{(-9.02)}}$ & $\text{77.74}_{\text{(-8.50)}}$ & $\text{89.25}_{\text{(-3.04)}}$\\
         & w/o Hori & $\text{79.65}_{\text{(-2.01)}}$ & $\text{84.20}_{\text{(-2.04)}}$ & $\text{91.90}_{\text{(-0.39)}}$ \\
         & w/o Vert & $\text{71.66}_{\text{(-10.00)}}$ & $\text{77.28}_{\text{(-8.96)}}$ & $\text{88.98}_{\text{(-3.31)}}$ \\
         \bottomrule
    \end{tabular}
    \caption{The ablation study of $\text{TIE}_{\text{MarkupLM}}$ on the development set of WebSRC. \dag denotes average results of 3 random seeds.}
    \label{tab:ablation}
\end{table}

To further investigate the contributions of key components, we make the following variants of TIE:
(1)\textbf{"w/o DOM"} means only using NPR graphs without the DOM trees.
(2)\textbf{"w/ ORD"} means using original sparse DOM trees instead of the denser version introduced in Sec.\ref{sec:model}.
(3)\textbf{"w/o NPR"} means only using the densified DOM trees without the NPR graphs.
(4)\textbf{"w/o Hori"} removes {\tt LEFT} and {\tt RIGHT} relations in NPR graph.
(5)\textbf{"w/o Vert"} removes {\tt UP} and {\tt DOWN} relations in NPR graph.

The results are shown in Table \ref{tab:ablation}, from which we have several observations and analysis:

First,
we investigate the contribution of DOM trees. The performance of \textbf{"w/o DOM"} drops slightly compared with original TIE, which indicates that the contributions of DOM trees are marginal. That may be because MarkupLM has leveraged XPaths to encode the logical information. Considering that XPaths are defined based on DOM trees, the information contained in XPaths and DOM trees may largely overlap.
Moreover, the results of \textbf{"w/ ORD"} show that densifying the DOM Tree is vitally important, as the original DOM tree is extremely sparse and will significantly lower the performance of TIE.

Finally, the NPR graphs have great contributions as the performance of \textbf{"w/o NPR"} drops significantly. It is because NPR graphs can help TIE efficiently model the informative relations such as {\tt key-value} and {\tt header-cell}, as they are often arranged in the same row or column.
Moreover, we further investigate the contribution of different relations in NPR graphs by \textbf{"w/o Hori"} and \textbf{"w/o Vert"}. Note that, we keep the number of parameters of TIE unchanged among these experiments, which means no {\tt horizontal} relations in NPR graphs will result in more attention heads assigned to {\tt vertical} relations.
The results show that, in WebSRC, {\tt vertical} relations are much more important than {\tt horizontal} relations. That is because most of the websites in WebSRC are constructed row-by-row, which means that the tags of {\tt horizontal} relations are often located near each other in the HTML codes while those of {\tt vertical} relations may be located far apart. Therefore, in most cases, the {\tt horizontal} relations are easier to capture in the context encoder without the help of NPR graph, while the {\tt vertical} relations can hardly achieve that.

\section{Related Work}

\paragraph{Question Answering~(QA)}

In recent years, a large number of QA datasets and tasks have been proposed, ranging from Plain text QA (i.e. MRC) \citep{rajpurkar-etal-2016-squad, joshi-etal-2017-triviaqa, lai-etal-2017-race, yang-etal-2018-hotpotqa, reddy-etal-2019-coqa} to QA over KB \citep{berant-etal-2013-semantic, bao-etal-2016-constraint, yih-etal-2016-value, talmor-berant-2018-web, lc}, Table QA \citep{pasupat-liang-2015-compositional, chen-etal-2020-hybridqa, ttqa}, Visual QA~(VQA) \citep{vqa, fvqa, okvqa}, and others. However,
the topological information in the textual inputs is either absent (plain text) or simple and explicitly provided (KB/tables). The QA task based on semi-structured HTML codes with implicit and flexible topology is under-researched.

Among these tasks, Table QA is the most similar to the Web-based SRC task, as there are many tables in the WebSRC dataset. To solve the problem,
\citet{dot} first selects candidate answer cells according to cell embeddings from the whole table and then finds the accurate answer cell from the candidates. Their method enables the model to handle larger tables at little cost.
On the other hand, \citet{glass-etal-2021-capturing} introduces row and column interactions into their models and determines the final answers based on the top-ranked relevant rows and columns.
In addition, Text-to-SQL is another group of methods to tackle Table QA problems and has been widely studied recently \citep{yu-etal-2018-typesql, bogin-etal-2019-representing, wang-etal-2020-rat, cao-etal-2021-lgesql, chen-etal-2021-decoupled, chen-etal-2021-shadowgnn, s2sql}. They use databases to store the source tables and translate natural language queries into Structured Query Language~(SQL) to retrieve answers from the databases.
It is worth noticing that these methods are highly coupled with the data format and requires 
simple and neat structures.
Therefore, their methods are not suitable for Web-based SRC tasks.

\paragraph{Web Question Answering}

Recent works which mentioned Web Question Answering mainly focus on the post-processing of the plain texts \citep{twqa2, twqa1} or tables \citep{zhang-etal-2020-graph} resulting from the searching engine.
Moreover, \citet{dslwqa} has tried to answer fixed-form questions based on raw HTML codes with the help of Domain-Specific Language~(DSL).
Apart from the above works, \citet{websrc} proposed a QA task called Web-Based SRC which is targeted at the comprehension of the structured web pages using raw HTML codes.
The method they proposed is to treat the HTML tags as special tokens and directly feed the raw flattened HTML codes into the PLM. They also tried to leverage screenshots as auxiliary information.
Later, \citet{markuplm} introduced a novel pre-trained model called MarkupLM specifically for XML-based documents. They adopted a new kind of position embedding generated from the XPath of each token to implicitly encode the logical information of XML codes.
In this work, we further explicitly introduce the topological structures to the models with the help of DOM trees and NPR graphs. A newly designed tag-level QA model with a two-stage pipeline is leveraged to take advantage of these graphs.

\section{Conclusion \& Future Work}

In this paper, we proposed a tag-level QA model called TIE to better understand the topological information contained in the structured web pages.
Our model explicitly captures two of the most informative topological structures of the web pages, logical and spatial structures, by DOM trees and NPR graphs, respectively.
With the proposed two-stage pipeline, 
we conduct extensive experiments on the WebSRC dataset.
Our TIE successfully achieves SOTA performances and the contributions of its key components are validated.

Although our TIE can achieve much high performance compared with traditional QA models on SRC tasks, more improvements are still needed. Specifically, as our two-stage system needs a separated token-level QA model to generate final answer spans, the parameter numbers and computation consumption will be at least doubled. We have tried to tackle this problem by sharing parameters between the context encoder and the token-level QA model used in the \textit{answer refining} stage. But the results are not promising. Therefore, we leave this problem for future work.
\section*{Acknowledgements}
We sincerely thank the anonymous reviewers for their valuable comments. This work has been supported by the China NSFC Projects (No. 62120106006 and No. 62106142), Shanghai Municipal Science and Technology Major Project (2021SHZDZX0102), CCF-Tencent Open Fund and Startup Fund for Youngman Research at SJTU (SFYR at SJTU).
\bibliography{anthology,custom}
\bibliographystyle{acl_natbib}

\appendix

\section{Detail Setup}
\label{sec:setup}
To train the model, we use AdamW \citep{adamw} with a linear schedule as our optimizer. As for the learning rate, we search for the best learning rate between 1e-6 and 5e-5.
Finally, TIE is trained and evaluated on four Nvidia A10 Graphics Cards with batch size 32 for two epochs.
Moreover, for BASE size models~(12 heads in total), we use DOM Trees to generate the mask matrix for 4 attention heads and each of the 4 NPR graphs for 2 attention heads. And for LARGE size models~(16 heads in total), we add one more attention head using each of the 4 NPR graphs.

\section{Additional Case Study} \label{sec:add}

\begin{figure*}[t]
    \centering
    \includegraphics[width=\hsize]{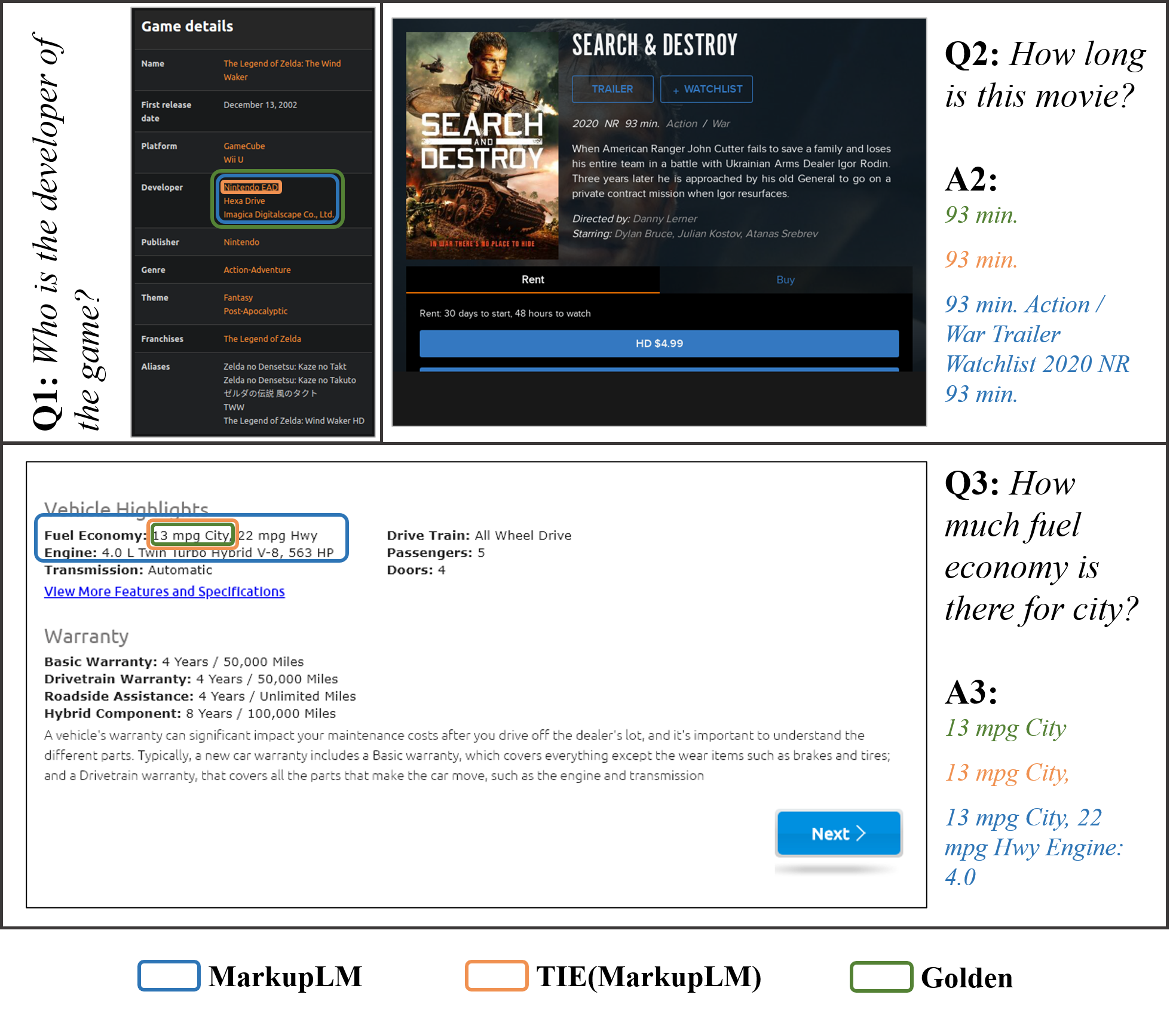}
    \caption{Examples of the results from \textit{KV} type websites in the development set.}
    \label{fig:comk}
\end{figure*}

\begin{figure*}[t]
    \centering
    \includegraphics[width=\hsize]{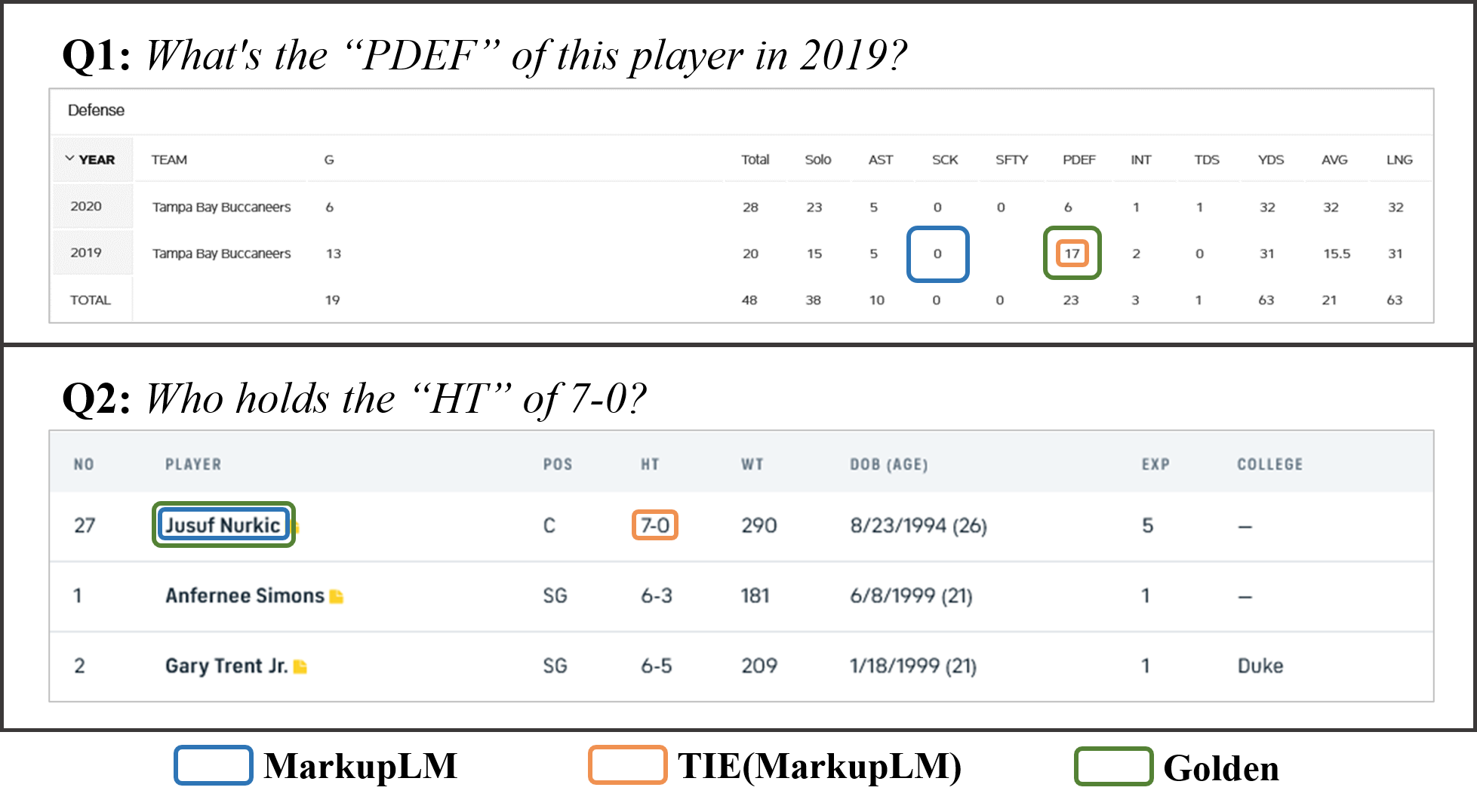}
    \caption{Examples of the results from \textit{Table} type websites in the development set.}
    \label{fig:comt}
\end{figure*}

\begin{figure*}[t]
    \centering
    \includegraphics[width=\hsize]{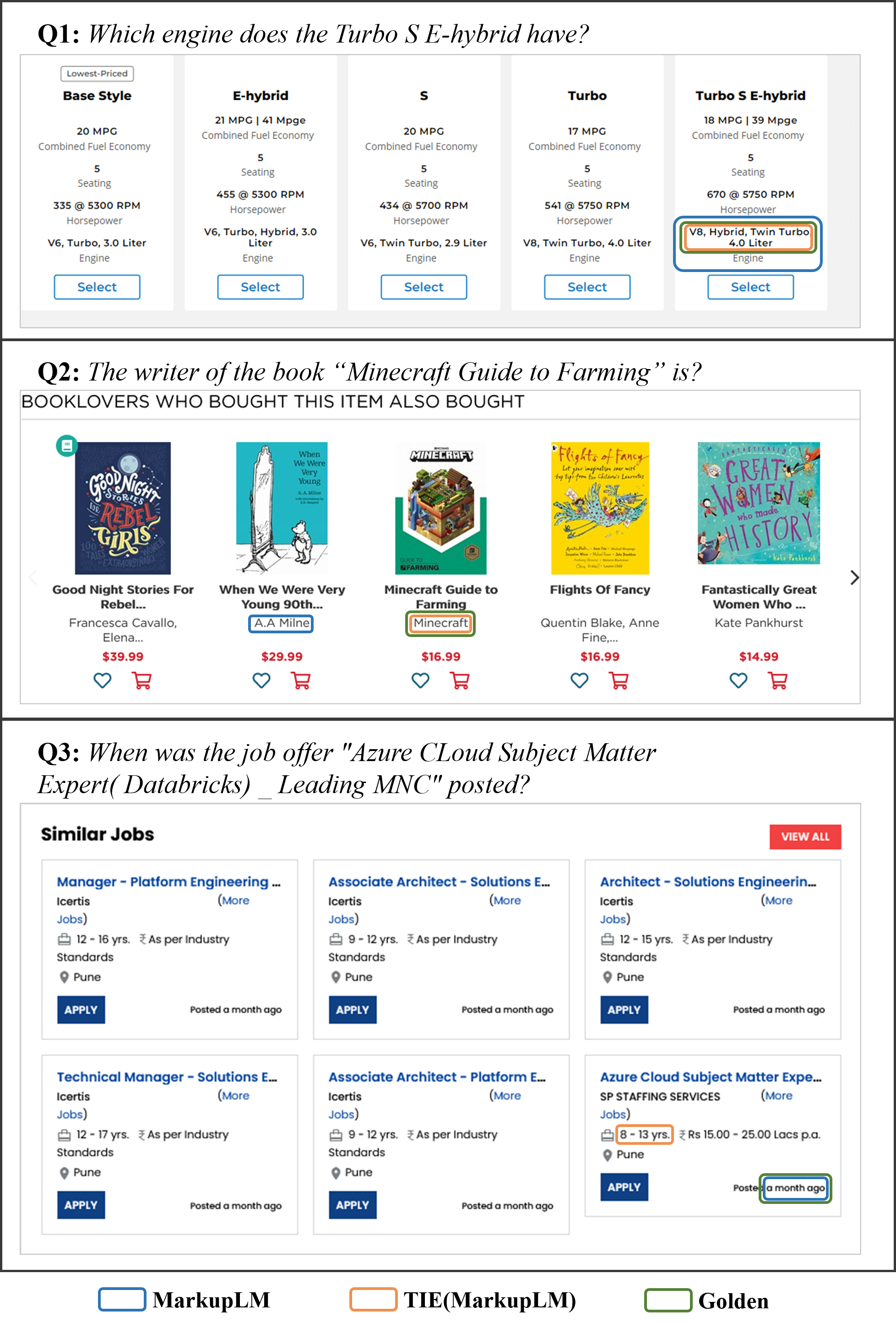}
    \caption{Examples of the results from \textit{Compare} type websites in the development set.}
    \label{fig:comc}
\end{figure*}

Figure \ref{fig:comk}, \ref{fig:comt}, and \ref{fig:comc} shows the typical examples of the QA pairs in \textit{KV}, \textit{Table}, and \textit{Compare} websites, respectively.

Through detailed analysis, we found that TIE can better capture the long-range relations which have obvious spacial relations, such as {\tt header-cell} and {\tt entity-attribute} (see Fig. \ref{fig:comk} \textbf{Q3}, Fig. \ref{fig:comt} \textbf{Q1}, and Fig. \ref{fig:comc} \textbf{Q2}). On the other hand, as TIE focuses more on tag-level structure understanding, its ability to understand token-level semantics may be weaker, which leads to some of the TIE's wrong predictions (see Fig. \ref{fig:comk} \textbf{Q1}, Fig. \ref{fig:comt} \textbf{Q2}, and Fig. \ref{fig:comc} \textbf{Q3}). In addition, TIE has a better awareness of tag boundaries, which has been proven useful when answering questions with blurry boundaries (see Fig. \ref{fig:comk} \textbf{Q2}, \textbf{Q3}, and Fig. \ref{fig:comc} \textbf{Q1}).

\end{document}